\documentclass[10pt,twocolumn,letterpaper]{article}

\usepackage[pagenumbers]{cvpr}      

\usepackage{cuted} 

\usepackage{currfile} 

\usepackage{caption} 
\definecolor{cvprblue}{rgb}{0.21,0.49,0.74}
\usepackage[pagebackref,breaklinks,colorlinks,allcolors=cvprblue]{hyperref}
\usepackage[table]{xcolor}
\usepackage{multirow}

\def\paperID{8318} 
\def\confName{CVPR}
\def\confYear{2027}

\title{Video2Asset: Online Construction of Persistent 3D Street-Scene Assets}

\author{Zihan Zheng$^{1}$$^*$
\and
Chunlei Cai$^{2}$
\and
Zhenlong Wu$^{1}$
\and
Linjun Li$^{2}$
\and 
Yichen Gong$^{2}$
\quad \quad
Xiaoyun Zhang$^{1}$
\quad \quad
Qiang Hu$^{1}$$^{\dagger}$
\\
$^{1}$ Cooperative Medianet Innovation Center, Shanghai Jiao Tong University \\ $^{2}$ Agentic Labs, Acrab AI \\
\vspace{-3em}
}

\begin{document}
\maketitle
{\let\thefootnote\relax\footnote{\
        ${ }^{*}$ Work done as a research intern at Acrab AI. 
        ${ }^{\dagger}$ The corresponding author is Qiang Hu(qiang.hu@sjtu.edu.cn)
}}\par

\begin{strip}
\centering
\vspace{-5em}
\includegraphics[width=\linewidth]{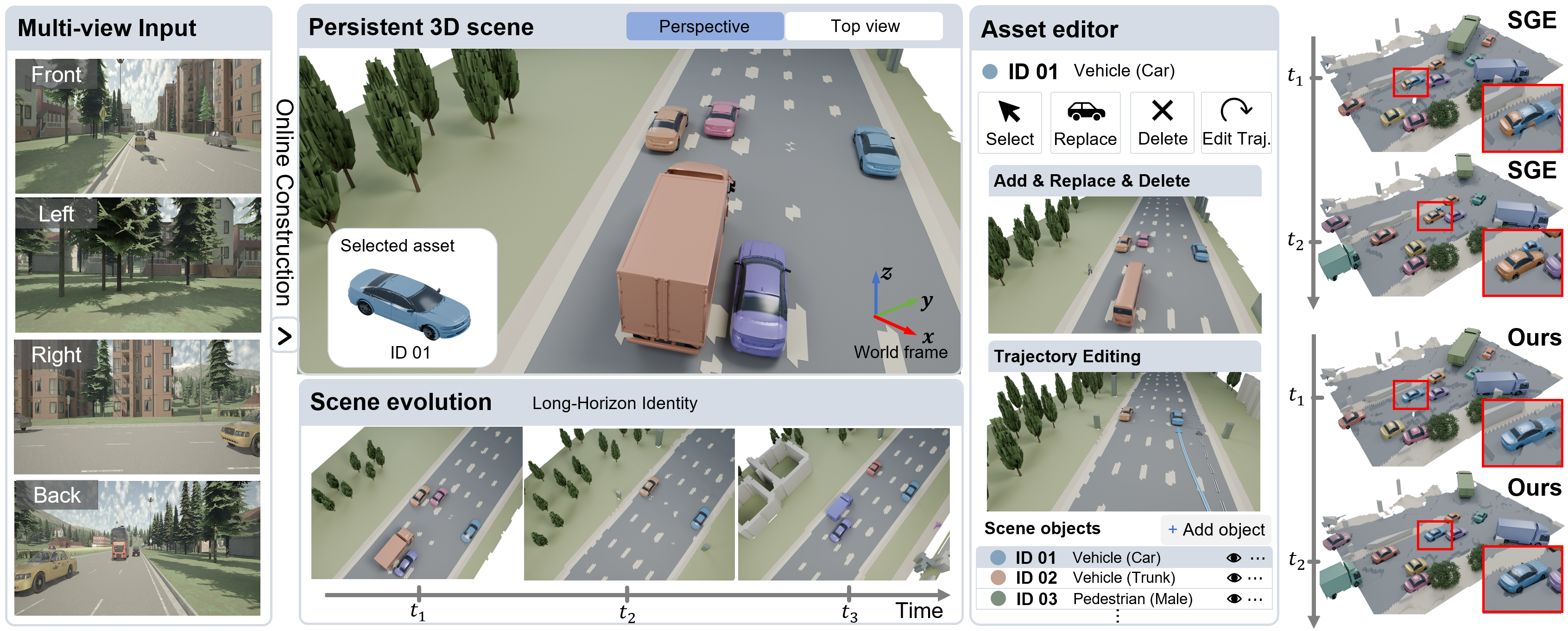}
\captionsetup{hypcap=false}
\vspace{-1.5em}
\captionof{figure}{\textbf{Persistent and editable street-scene assets from video.} Left: Video2Asset converts multi-view video into an evolving 3D scene with persistent object identities, supporting mesh replacement, object deletion, and trajectory editing. Right: qualitative comparison with SGE. Our method maintains more consistent asset identities across frames.}
\captionsetup{hypcap=true}
\label{fig:teaser}
\end{strip}

\begin{abstract}
Online 3D street-scene asset construction turns multi-view video into a continually updated scene of independently editable objects.
The central challenge is to accumulate partial, intermittent observations under motion and occlusion without fragmenting identities or mixing evidence across objects.
To address this challenge, we propose Video2Asset, an online framework for constructing persistent 3D assets from multi-view video.
Using occupancy as an intermediate representation, we construct identity-aware observations with Historical Carrier Guidance for instance decoding and an identity encoder fusing instance-aligned 3D features with multi-view appearance.
Accumulated-Evidence Association compares current observations with identity, geometric, and appearance evidence accumulated within asset records to determine identity inheritance and update world-coordinate states.
We instantiate meshes from an asset library using these states and combine them with an occupancy-derived static background to support scene visualization, mesh replacement, object deletion, and trajectory editing.
Experiments on several datasets demonstrate improvements in temporal identity consistency and asset quality across multiple occupancy models, including a 15.4\% reduction in asset fragmentation on nuScenes with SGE.
\end{abstract}
    
\section{Introduction}
\label{sec:intro}

Editable 3D street scenes allow individual objects to be moved, replaced, or removed within a shared environment.
Constructing such scenes online from multi-view video requires capturing object-level geometry and semantics, then accumulating observations into assets with persistent identities and evolving world-coordinate states.
The central challenge is to accumulate partial, intermittent observations under motion and occlusion without fragmenting one object across assets or mixing evidence from distinct objects.

Neural radiance fields (NeRFs) and 3D Gaussian Splatting (3DGS) reconstruct street scenes by fitting multi-view images, with object-level formulations supporting independent manipulation of scene elements~\cite{ost2021neuralscenegraphs,yan2024street,chen2025omnire}.
Under sparse view coverage, however, fitting image colors alone can reproduce observed appearance without adequately constraining the underlying 3D structure~\cite{niemeyer2022regnerf,zhu2024fsgs}.
Occupancy representations trained with explicit 3D geometric and semantic supervision encode occupied regions and scene categories~\cite{wei2023surroundocc,zhang2023occformer}, providing a structured spatial basis for street-scene construction.

Temporal occupancy models propagate scene evidence across frames~\cite{yu2026superocc,zuo2025gaussianworld}, while 4D panoptic occupancy tracking couples geometry and semantics with cross-frame instance identities~\cite{chen2025trackocc,luz2026lags,luz2026sge}.
These capabilities provide the temporal perception and association basis for accumulating observations into persistent assets.
However, under occlusion and instance competition, query-based identity propagation suffers from identity fragmentation and erroneous instance merging, disrupting the object histories required for consistent asset construction.
Turning these predictions into editable assets further requires consolidating observations into persistent object records with evolving world-coordinate states.

To address these challenges, we propose Video2Asset, an online framework that uses occupancy as an intermediate representation to construct persistent 3D street-scene assets from multi-view video (Figure~\ref{fig:teaser}).
First, we construct identity-aware observations that capture object geometry and semantics while providing identity evidence for cross-frame association.
Historical Carrier Guidance (HCG) uses motion-aligned historical instance regions to guide current feature aggregation, providing object-specific spatial support when observations are incomplete.
An identity encoder fuses instance-aligned 3D features with multi-view appearance to provide identity evidence for asset association.

Second, Accumulated-Evidence Association (AEA) links observations to persistent assets using evidence accumulated within their records.
It constructs observation--asset relation features by comparing the observation with each candidate's identity, geometric, and appearance evidence.
Pairwise scores rank the candidates, while a separate inheritance predictor reads the full candidate relation set to determine whether to inherit an existing identity.
One-to-one assignment extends matched asset histories and updates world-coordinate states, while unassigned observations initialize new records.
We instantiate meshes from an asset library using these states and combine them with an occupancy-derived static background to support object deletion, trajectory editing, and mesh replacement.

Our contributions are summarized as follows:
\begin{itemize}
\item We propose Video2Asset, an online framework that turns multi-view video into persistent 3D street-scene assets with world-coordinate states and independently editable library-mesh instances.
\item We develop identity-aware observation construction that combines HCG-guided instance decoding with an identity encoder fusing instance-aligned 3D features and multi-view appearance.
\item We introduce AEA, which compares current observations with identity, geometric, and appearance evidence accumulated within asset records for candidate ranking and identity-inheritance decisions.
\end{itemize}

\section{Related Work}
\label{sec:related-work}

\textbf{Appearance-based scene construction.}
Radiance-field and Gaussian methods reconstruct static or dynamic scenes by fitting multi-view images~\cite{mildenhall2020nerf,kerbl2023gaussiansplatting,zheng2024jointrf,zheng2024hpc,hu2025vrvvc,hu2025_4dgc,zheng2025_4dgcpro,zheng2026cdgs}.
Object-level formulations organize these representations for editing and simulation~\cite{ost2021neuralscenegraphs,yan2024street,chen2025omnire,zhou2024hugsim}.
Under sparse views, fitting observed colors alone can overfit appearance while leaving geometry underconstrained~\cite{niemeyer2022regnerf,zhu2024fsgs}, motivating geometric guidance~\cite{zhong2025prismgs,hu2026gensplatcodec}.
We use occupancy as an intermediate representation, grounding identity-aware observations in geometric and semantic supervision to capture scene structure for asset construction.

\textbf{Geometry-based scene construction.}
Point-cloud and surface reconstruction recover scene geometry through multi-view depth estimation or implicit surface modeling~\cite{yao2018mvsnet,wang2021neus,li2023neuralangelo}.
Semantic scene completion extends this spatial modeling to jointly infer occupancy and categories in visible and occluded regions~\cite{cao2022monoscene,li2023voxformer}.
For surrounding scenes, occupancy prediction encodes geometric and semantic structure with voxel grids, tri-perspective features, or semantic Gaussians~\cite{wei2023surroundocc,zhang2023occformer,huang2023tpvformer,huang2024gaussianformer}.
Panoptic formulations further distinguish individual instances within occupied regions~\cite{wang2024panoocc,liu2024sparseocc}.
We extend this spatial modeling across time, preserving object identity continuity and consolidating observations into persistent assets with evolving world-coordinate states.

\textbf{Temporal scene and asset modeling.}
Temporal occupancy models integrate cross-frame evidence through propagated scene representations or spatiotemporal memory~\cite{zuo2025gaussianworld,yu2026superocc,leng2025stocc}.
TrackOcc, LaGS, and SGE make identities explicit through propagated instance queries~\cite{chen2025trackocc,luz2026lags,luz2026sge}.
Under occlusion and instance competition, identity fragmentation and erroneous instance merging disrupt the object histories needed for asset construction.
History-based tracking accumulates object evidence across observations~\cite{cai2022memot,gao2023memotr,lee2025lomm}, while object-level mapping maintains evolving states in world coordinates~\cite{schmid2024khronos}.
We selectively use accumulated identity, geometric, and appearance evidence to associate current observations with persistent assets.

\section{Method}
\label{sec:method}

\begin{figure*}[t]
    \centering
    \includegraphics[width=\linewidth]{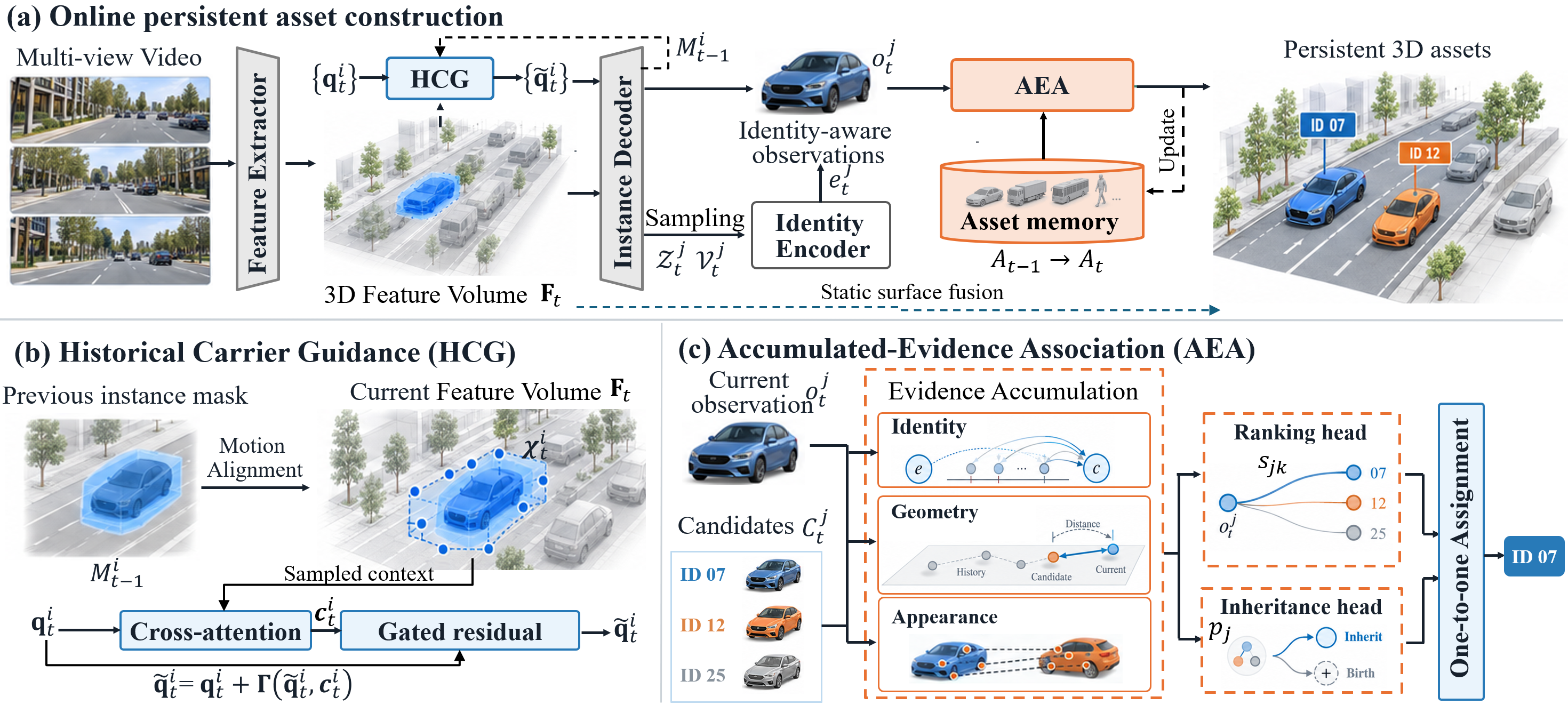}
    \vspace{-1.5em}
    \caption{\textbf{Overview of Video2Asset.} (a) Multi-view video is converted into identity-aware observations through occupancy modeling. The observations are associated with persistent asset records to update world-coordinate states and construct an editable 3D scene. (b) Historical Carrier Guidance (HCG) aligns previous instance masks to guide query-conditioned aggregation of current occupancy features. (c) Accumulated-Evidence Association (AEA) compares current observations with identity, geometric, and appearance evidence accumulated in candidate asset records for matching and identity-inheritance prediction, followed by one-to-one assignment.}
    \label{fig:method}
    \vspace{-1em}
\end{figure*} 

As illustrated in Figure~\ref{fig:method}(a), Video2Asset constructs persistent, editable 3D street scenes online from multi-view video.
We first construct identity-aware observations using occupancy as an intermediate representation: Historical Carrier Guidance guides current feature aggregation with aligned previous instance regions, and an identity encoder summarizes instance-aligned evidence for cross-frame comparison (Section~\ref{sec:method:observations}).
Accumulated-Evidence Association then relates current observations to accumulated asset evidence to determine identity inheritance and update persistent records (Section~\ref{sec:method:assets}).
The associated observations update world-coordinate asset states, from which object meshes are instantiated and composed with the static background.

\subsection{Problem Formulation}
\label{sec:method:formulation}

Given a multi-view video stream of a street scene, we construct a 3D scene representation at time $t$ using only the calibrated images $\mathcal I_{1:t}$ and camera poses $\mathcal T_{1:t}$ available up to that time.
The output is $\mathcal W_t=(\mathcal S_t,\mathcal A_t)$, where $\mathcal S_t$ denotes the static scene background and $\mathcal A_t=\{A_t^k\}_k$ the set of maintained object assets.

The goal is to consolidate partial, intermittent observations of the same physical object into a persistent asset record and update its world-coordinate state.
For asset $k$, the online state is
\begin{equation}
 A_t^k=(\iota^k,c^k,\mathcal H_t^k,\mathcal G_t^k),
\label{eq:method:asset}
\end{equation}
where $\iota^k$ is a persistent ID and $c^k$ the semantic class.
The history $\mathcal H_t^k$ accumulates identity descriptors, multi-view appearance features, and timestamped centers from associated observations, while $\mathcal G_t^k$ describes the asset's spatial extent, world position, and motion.
World position, scale, and orientation determine the placement of a library mesh, indexed by $\iota^k$ for object-level editing and replacement.

At each time step, $\mathcal O_t=\{o_t^j\}_j$ denotes the identity-aware observations constructed from the video.
Object-asset construction proceeds through observation formation and asset-state updates:
\begin{equation}
\begin{aligned}
 (\mathcal O_t,h_t) &= F_{\mathrm{obs}}(\mathcal I_t,\mathcal T_t,h_{t-1}),\\
 \mathcal A_t &= F_{\mathrm{asset}}(\mathcal O_t,\mathcal A_{t-1}).
\end{aligned}
\label{eq:method:decomposition}
\end{equation}
Here $F_{\mathrm{obs}}$ constructs $\mathcal O_t$ and updates $h_t$, the temporal perception state containing retained instance support.
$F_{\mathrm{asset}}$ associates current observations with maintained records, updating their histories and spatial states or creating new assets.
The static background $\mathcal S_t$ is constructed from predicted occupancy in world coordinates.

\subsection{Identity-aware Observation Construction}
\label{sec:method:observations}

Viewpoint changes, object motion, and occlusion continually alter the visible region of an object, leaving current-frame evidence incomplete.
The observation mapping $F_{\mathrm{obs}}$ builds $\mathcal O_t$ with instance-level spatial support and identity descriptors that can be compared across frames.
Occupancy modeling~\cite{chen2025trackocc,luz2026lags,luz2026sge} provides a 3D feature volume $\mathbf F_t$ from $\mathcal I_t$.
An instance decoder uses a set of queries $\{\mathbf q_t^i\}_i$ as instance hypotheses to read these features and predict instance masks and attributes.

\textbf{Historical Carrier Guidance.}
HCG uses previous instance masks retained in $h_{t-1}$ as query-specific spatial carriers to determine where to aggregate current occupancy features before instance decoding (Figure~\ref{fig:method}(b)).
For a query $\mathbf q_t^i$ with valid historical support, let $M_{t-1}^i$ denote the previous frame's final instance mask linked to it through the source-query correspondence.
Points sampled from this mask are transformed into the current frame using camera poses $\mathcal T_{t-1:t}$ and the available object-velocity estimate, yielding the aligned spatial support $\mathcal X_t^{i}$.
Attention conditioned on $\mathbf q_t^i$ aggregates current occupancy features at these locations into a context $\mathbf c_t^{i}$, and a gated residual updates the corresponding query:
\begin{equation}
\begin{aligned}
 \mathbf c_t^{i} &= \operatorname{Attn}\!\left(
 \mathbf q_t^{i},\{\mathbf F_t(\mathbf x):\mathbf x\in\mathcal X_t^{i}\}\right),\\
 \widetilde{\mathbf q}_t^{i} &= \mathbf q_t^{i}
 +\Gamma(\mathbf q_t^{i},\mathbf c_t^{i}).
\end{aligned}
\label{eq:method:guided-query}
\end{equation}
Here $\Gamma$ denotes the gated residual mapping.
The updated queries are decoded into current instance masks, semantic classes, and confidence scores.

\textbf{Instance-aligned identity encoding.}
For each decoded instance $j$, the final mask $M_t^j$ provides a common spatial support for geometric measurements and identity evidence from occupancy features and multi-view appearance.
Occupancy features sampled within this region are combined with relative coordinates, mask confidence, and visibility to form spatial tokens $\mathcal Z_t^j$.
Valid appearance tokens $\mathcal V_t^j$ are extracted from image features in the corresponding multi-view projection regions.
The identity encoder uses $\mathcal Z_t^j$ as its primary representation and incorporates $\mathcal V_t^j$ through a cross-attention residual.
The fused evidence is aggregated into a normalized identity descriptor:
\begin{equation}
 \mathbf e_t^j=
 \mathcal N\!\left(E_{\mathrm{id}}(\mathcal Z_t^j,\mathcal V_t^j)\right),
\label{eq:method:identity}
\end{equation}
where $E_{\mathrm{id}}$ is the identity encoder that fuses and aggregates the tokens, $\mathcal N(\cdot)$ denotes $L_2$ normalization, and $\mathbf e_t^j$ is the normalized identity descriptor.

The identity descriptor and instance measurements together form each observation $o_t^j\in\mathcal O_t$:
\begin{equation}
 o_t^j=(M_t^j,c_t^j,s_t^j,\mathbf g_t^j,\mathbf e_t^j,
                  \mathcal Z_t^j,\mathcal V_t^j,\pi_t^j).
\label{eq:method:observation}
\end{equation}
Here $c_t^j$ and $s_t^j$ are the predicted class and confidence, and $\mathbf g_t^j$ contains the instance center, spatial extent, and motion measurements.
The source-query index $\pi_t^j$ links the final mask to its generating query; AEA in Section~\ref{sec:method:assets} assigns the persistent asset ID.
The final masks and source-query correspondences are retained in $h_t$ for subsequent observation construction.

\textbf{Identity learning.}
The identity encoder is learned from predicted instance evidence using earlier observations from the same continuous scene segment: the same physical instance provides positives, and other instances of the same class provide negatives.
The identity objective combines four terms:
\begin{equation}
\begin{aligned}
 \mathcal L_{\mathrm{id}}={}&
 \mathcal L_{\mathrm{contrast}}+\mathcal L_{\mathrm{attract}}+\mathcal L_{\mathrm{rank}}+\mathcal L_{\mathrm{negative}}.
\end{aligned}
\label{eq:method:identity-loss}
\end{equation}
Here $\mathcal L_{\mathrm{contrast}}$ maximizes the summed probability mass of historical positives among positive and same-class negative observations, while $\mathcal L_{\mathrm{attract}}$ increases mean positive similarity.
$\mathcal L_{\mathrm{rank}}$ enforces a margin between the least similar positive and the most similar negative, and $\mathcal L_{\mathrm{negative}}$ penalizes the maximum negative similarity above a margin.
Only terms supported by the available positive and negative samples contribute to the objective.

\subsection{Association and Persistent Asset Construction}
\label{sec:method:assets}

The asset-update mapping $F_{\mathrm{asset}}$ links $\mathcal O_t$ to records in $\mathcal A_{t-1}$ using their accumulated evidence.

\textbf{Accumulated-Evidence Association.}
For each $o_t^j$, AEA retrieves candidate assets using the minimum BEV distance from the current observation center to each asset's previously observed centers, expressed in the current ego frame.
Their histories $\mathcal H_{t-1}^k$ form the candidate collection $\mathcal C_t^j$, with multiple observations retained separately for each candidate asset.
AEA compares the observation with identity, geometric, and appearance evidence accumulated within each candidate, using their respective history-reading and comparison mechanisms to construct pairwise relation features (Figure~\ref{fig:method}(c)).
For an observation--asset pair, $\ell$ indexes the past observations of candidate asset $k$.

\emph{Identity evidence.}
Each candidate asset retains a separate identity descriptor $\mathbf e_{k\ell}$ for each associated observation, accumulating evidence across time and visibility conditions.
Using the current descriptor $\mathbf e_t^j$ as the attention query, we obtain an observation-conditioned identity summary:
\begin{equation}
 \bar{\mathbf e}_{jk}
 =\operatorname{Attn}\!\left(
 \mathbf e_t^j,\{\mathbf e_{k\ell}\}_{\ell}\right).
\label{eq:method:identity-history}
\end{equation}
Only valid history entries participate in this candidate-specific read, allowing the same asset to yield different summaries for different observations.
The current descriptor and historical summary $\bar{\mathbf e}_{jk}$ provide identity evidence from observations and memory, while their difference captures the discrepancy between them.
These terms are combined with the pair's spatial--temporal and observation context to form the identity-based relation feature $\mathbf r^{\mathrm{id}}$.

\emph{Geometric evidence.}
Historical centers are transformed from world coordinates into the current ego frame and compared with the current observation center in the BEV plane.
Relative offsets are measured from each historical center to the current center.
We collect minimum Euclidean distances over recent history windows, the mean offset and its magnitude over all valid records, and the offset and age of the spatially closest record.
Together with overall and recent valid-history coverage, these statistics form the geometry-based relation feature $\mathbf r^{\mathrm{geo}}$.

\emph{Appearance evidence.}
Appearance evidence is compared across views before it is aggregated across history.
For historical observation $\ell$, $S_\ell(v,w)$ is the cosine similarity between separately projected, $L_2$-normalized features of current view $v$ and historical view $w$.
Each view finds its best match among valid views on the other side; averaging the two directional results gives matching strength $m_\ell$:
\begin{equation}
\begin{aligned}
 m_\ell
 &= \frac12\Bigl(
 \operatorname{mean}_{v}\max_{w}S_\ell(v,w)\\
 &\qquad +\operatorname{mean}_{w}\max_{v}S_\ell(v,w)
 \Bigr).
\end{aligned}
\label{eq:method:view-matching}
\end{equation}
Both means and maxima operate on valid views.
An MLP jointly encodes this matching strength, projected image-feature support, and valid-view fractions for both observations into a history relation feature $\mathbf z_\ell$.
Normalized weights $\beta_\ell$ predicted from these features aggregate history entries with valid views on both sides.
The aggregate is concatenated with cross-history matching and coverage summaries $\mathbf b$ to form the appearance-based relation feature:
\begin{equation}
 \mathbf r^{\mathrm{app}}=
 \Bigl[\sum_\ell\beta_\ell\mathbf z_\ell,\mathbf b\Bigr].
\label{eq:method:appearance-relation}
\end{equation}

These relation features support two complementary decisions: ranking candidate assets and deciding whether to inherit an existing identity.
For observation $j$ and candidate $k$, each relation branch $b$ predicts a pairwise matching score.
A separate inheritance head in each branch uses observation-conditioned attention over the full set of valid candidate relation features derived from $\mathcal C_t^j$ to produce $\mathrm{Birth}/\mathrm{Inherit}$ logits $\boldsymbol\lambda_j^b$.
The branch outputs are summed to obtain the observation--asset matching score $s_{jk}$ and the probability $p_j$ that observation $j$ inherits an existing identity:
\begin{equation}
\begin{aligned}
 s_{jk} &= \sum_b f_b(\mathbf r_{jk}^b),\\
 p_j &= \operatorname{softmax}\!\left(
 \sum_b\boldsymbol\lambda_j^b\right)_{\mathrm{Inherit}}.
\end{aligned}
\label{eq:method:association}
\end{equation}
Here $b\in\{\mathrm{id},\mathrm{geo},\mathrm{app}\}$, $\mathbf r_{jk}^b$ is the corresponding pairwise relation feature, and $f_b$ is its scoring MLP.

At inference, asset memory starts empty.
The $p_j$-based $\mathrm{Birth}/\mathrm{Inherit}$ decision selects observations eligible for identity inheritance.
Their candidate edges are greedily assigned in descending $s_{jk}$, with each observation and asset matched at most once per frame.
Matched observations retain the corresponding asset IDs $\iota^k$, while unassigned observations initialize new assets.

\begin{table*}[t]
    \centering
    \caption{\textbf{Quantitative comparison on CarlaOcc, nuScenes, and Waymo.} Shaded rows denote Video2Asset with the indicated occupancy model. Metrics cover observation quality, temporal consistency, and asset quality.}
    \vspace{-1em}
    \label{tab:exp:comparison}
    \small
    \setlength{\tabcolsep}{3pt}
    \renewcommand{\arraystretch}{1.08}
    \resizebox{\textwidth}{!}{%
    \begin{tabular}{cl*{11}{r}}
        \toprule
        \multirow[c]{2}{*}{Dataset} & \multirow[c]{2}{*}{Method} & \multicolumn{4}{c}{Observation Quality} & \multicolumn{4}{c}{Temporal Metrics} & \multicolumn{3}{c}{Asset Quality} \\
        \cmidrule(lr){3-6} \cmidrule(lr){7-10} \cmidrule(lr){11-13}
        & & mIoU $\uparrow$ & $\mathrm{STQ}_1\uparrow$ & $\mathrm{AQ}_1\uparrow$ & IoU $\uparrow$ & STQ $\uparrow$ & AQ $\uparrow$ & HOTA $\uparrow$ & IDF1 $\uparrow$ & $\mathrm{AP}_{25}\uparrow$ & $\mathrm{AP}_{50}\uparrow$ & $\mathrm{AF}\downarrow$ \\
        \midrule
        & TrackOcc~\cite{chen2025trackocc} & 33.92 & 25.91 & 19.80 & 56.91 & 12.89 & 4.90 & 11.25 & 7.24 & 8.88 & 6.62 & 6.35 \\
        \rowcolor{gray!10}
        \cellcolor{white} & \textbf{Ours \textit{(with TrackOcc)}} & \textbf{35.73} & \textbf{29.17} & \textbf{23.83} & \textbf{57.08} & \textbf{23.24} & \textbf{15.12} & \textbf{25.99} & \textbf{24.59} & \textbf{19.29} & \textbf{15.45} & \textbf{2.75} \\
        \addlinespace[2pt]
        & LaGS~\cite{luz2026lags} & 45.23 & 41.42 & 37.93 & 58.78 & 38.41 & 32.62 & 40.23 & 41.95 & 43.74 & 30.03 & 0.27 \\
        \rowcolor{gray!10}
        \cellcolor{white} & \textbf{Ours \textit{(with LaGS)}} & \textbf{45.36} & \textbf{42.13} & \textbf{39.13} & \textbf{58.81} & \textbf{38.63} & \textbf{32.90} & \textbf{40.36} & \textbf{42.19} & \textbf{45.24} & \textbf{31.17} & \textbf{0.25} \\
        \addlinespace[2pt]
        & SGE~\cite{luz2026sge} & 44.23 & 40.75 & 37.54 & 58.05 & 37.99 & 32.63 & 40.25 & 41.73 & 41.91 & 26.31 & 0.34 \\
        \rowcolor{gray!10}
        \cellcolor{white} \multirow{-6}{*}[2pt]{\rotatebox[origin=c]{90}{CarlaOcc}} & \textbf{Ours \textit{(with SGE)}} & \textbf{44.38} & \textbf{40.91} & \textbf{37.73} & \textbf{58.06} & \textbf{38.08} & \textbf{32.68} & \textbf{40.39} & \textbf{42.01} & \textbf{42.54} & \textbf{27.29} & \textbf{0.32} \\
        \midrule
        & TrackOcc~\cite{chen2025trackocc} & 33.64 & 20.51 & 12.50 & 63.37 & 13.14 & 5.14 & 13.58 & 10.76 & 7.54 & 1.27 & 2.34 \\
        \rowcolor{gray!10}
        \cellcolor{white} & \textbf{Ours \textit{(with TrackOcc)}} & \textbf{33.85} & \textbf{22.82} & \textbf{15.38} & \textbf{63.69} & \textbf{17.12} & \textbf{8.66} & \textbf{22.23} & \textbf{21.31} & \textbf{11.02} & \textbf{2.28} & \textbf{1.45} \\
        \addlinespace[2pt]
        & LaGS~\cite{luz2026lags} & 40.67 & 36.34 & 32.47 & 67.31 & 32.24 & 25.55 & 37.16 & 41.70 & 36.88 & 24.53 & 0.26 \\
        \rowcolor{gray!10}
        \cellcolor{white} & \textbf{Ours \textit{(with LaGS)}} & \textbf{40.84} & \textbf{36.47} & \textbf{32.56} & \textbf{67.32} & \textbf{32.47} & \textbf{25.82} & \textbf{37.24} & \textbf{41.74} & \textbf{37.08} & \textbf{24.56} & \textbf{0.24} \\
        \addlinespace[2pt]
        & SGE~\cite{luz2026sge} & 43.36 & 38.45 & 34.10 & 71.66 & 34.30 & 27.14 & 38.79 & 44.19 & 38.26 & 25.14 & 0.26 \\
        \rowcolor{gray!10}
        \cellcolor{white} \multirow{-6}{*}[2pt]{\rotatebox[origin=c]{90}{nuScenes}} & \textbf{Ours \textit{(with SGE)}} & \textbf{43.56} & \textbf{38.64} & \textbf{34.29} & \textbf{71.69} & \textbf{34.51} & \textbf{27.35} & \textbf{38.92} & \textbf{44.46} & \textbf{38.46} & \textbf{25.20} & \textbf{0.22} \\
        \midrule
        & TrackOcc~\cite{chen2025trackocc} & 21.56 & 18.13 & 15.24 & 60.14 & 15.29 & 10.84 & 20.25 & 16.39 & 6.63 & 1.93 & 1.47 \\
        \rowcolor{gray!10}
        \cellcolor{white} & \textbf{Ours \textit{(with TrackOcc)}} & \textbf{21.56} & \textbf{18.54} & \textbf{15.94} & \textbf{60.15} & \textbf{15.82} & \textbf{11.61} & \textbf{22.67} & \textbf{19.37} & \textbf{8.39} & \textbf{2.52} & \textbf{1.04} \\
        \addlinespace[2pt]
        & LaGS~\cite{luz2026lags} & 24.48 & 23.47 & 22.51 & 61.47 & 21.14 & 18.25 & 30.84 & 30.07 & 27.89 & 13.54 & 0.28 \\
        \rowcolor{gray!10}
        \cellcolor{white} & \textbf{Ours \textit{(with LaGS)}} & \textbf{24.52} & \textbf{23.58} & \textbf{22.68} & \textbf{61.48} & \textbf{21.25} & \textbf{18.44} & \textbf{31.11} & \textbf{30.31} & \textbf{28.15} & \textbf{13.56} & \textbf{0.27} \\
        \addlinespace[2pt]
        & SGE~\cite{luz2026sge} & 25.25 & 24.09 & 22.99 & 61.86 & 21.87 & 18.95 & 31.91 & 31.06 & 27.58 & 13.87 & 0.25 \\
        \rowcolor{gray!10}
        \cellcolor{white} \multirow{-6}{*}[2pt]{\rotatebox[origin=c]{90}{Waymo}} & \textbf{Ours \textit{(with SGE)}} & \textbf{25.32} & \textbf{24.26} & \textbf{23.25} & \textbf{61.88} & \textbf{22.04} & \textbf{19.18} & \textbf{32.07} & \textbf{31.31} & \textbf{28.26} & \textbf{14.00} & \textbf{0.23} \\
        \bottomrule
    \end{tabular}
    }
    \vspace{-1em}
\end{table*}

\begin{table}[t]
    \centering
    \caption{\textbf{Asset fragmentation under low visibility.} Experiments are conducted on nuScenes dataset.}
    \vspace{-1em}
    \label{tab:exp:low-visibility}
    \small
    \setlength{\tabcolsep}{4pt}
    \renewcommand{\arraystretch}{1.0}
    \begin{tabular}{p{\dimexpr0.28\columnwidth-2\tabcolsep\relax} *{3}{>{\raggedleft\arraybackslash}p{\dimexpr0.24\columnwidth-2\tabcolsep\relax}}}
        \toprule
        AF $\downarrow$ & TrackOcc & LaGS & SGE \\
        \midrule
        Baseline & 2.48 & 0.87 & 0.95 \\
        \textbf{Ours} & \textbf{1.90} & \textbf{0.69} & \textbf{0.86} \\
        \bottomrule
    \end{tabular}
    \vspace{-1.5em}
\end{table}

\textbf{Association learning.}
Association training follows identity learning, beginning with histories built from predicted observations under ground-truth assignments before continuing with model-controlled rollouts.
During rollout, predicted associations update the histories used by subsequent decisions, with state retained across training chunks.
Ground truth supplies supervision targets; observation features and geometry remain predicted.

Association supervision uses two branch-specific objectives:
\begin{equation}
 \mathcal L_{\mathrm{assoc}}=
 \mathcal L_{\mathrm{edge}}+\mathcal L_{\mathrm{exist}}.
\label{eq:method:association-loss}
\end{equation}
The candidate-scoring and identity-inheritance parameters are optimized separately by $\mathcal L_{\mathrm{edge}}$ and $\mathcal L_{\mathrm{exist}}$, respectively.
The former applies BCE to positive edges and the hardest negative edge; the latter uses two-class cross-entropy over $\mathrm{Birth}$ and $\mathrm{Inherit}$.

\textbf{Persistent asset construction.}
After identity assignment, each $o_t^j$ updates $\mathcal H_t^k$ and $\mathcal G_t^k$ of its asset $A_t^k$, with records $\mathcal A_t$ maintained across frames within each independently processed sequence.
Unmatched assets retain their stored states and evidence, while velocity-extrapolated centers of dynamic assets provide association context. Each asset uses a library mesh selected by $c^k$, with position and scale determined by $\mathcal G_t^k$ and observation geometry, and orientation informed by motion direction. The static background $\mathcal S_t$ combines a ground surface causally fused from predicted road, sidewalk, and terrain occupancy in world coordinates, smoothed surfaces of predicted building occupancy, and vegetation meshes instantiated from the corresponding predicted semantic regions.
The background and object meshes form $\mathcal W_t$, with persistent asset IDs indexing object meshes for deletion, trajectory editing, and mesh replacement.

\section{Experiments}
\label{sec:experiments}

\newcommand{\experimentplaceholder}[2]{\fbox{\parbox[c][#1][c]{0.94\linewidth}{\centering\small #2}}}

\subsection{Configurations}
\label{sec:experiments:configurations}

\textbf{Datasets.}
We evaluate Video2Asset on three multi-view driving datasets: CarlaOcc, nuScenes, and Waymo.
CarlaOcc~\cite{feng2026carlaocc} comprises 104 synthetic sequences, each with 1,000 frames captured by six surround-view cameras at 20\,Hz, and provides  panoptic occupancy labels.
nuScenes~\cite{caesar2020nuscenes} comprises 1,000 real-world scenes, each lasting 20 seconds, with six surround-view cameras and keyframe annotations at 2\,Hz.
Waymo~\cite{sun2020waymo} comprises 1,150 real-world scenes, each lasting 20 seconds, with five cameras operating at 10\,Hz.
Occupancy annotations for nuScenes and Waymo are provided by Occ3D~\cite{tian2023occ3d}.

\textbf{Implementation.}
Experiments are conducted on NVIDIA A800 GPUs.
We follow the original training procedures of TrackOcc, LaGS, and SGE.
The image backbones are ResNet-50 for TrackOcc and VoVNetV2-99 for LaGS and SGE.
After occupancy and HCG training, we freeze perception to train the identity encoder, then freeze observation construction to train AEA.
Inference proceeds online, retaining observation histories and world-coordinate asset states across frames and resetting asset memory at the start of each independent sequence.

\textbf{Metrics.}
We evaluate our method along three dimensions: observation quality, temporal consistency, and asset quality.
Observation quality assesses per-frame geometry, semantics, and instance grouping using mean Intersection-over-Union (mIoU), binary occupancy Intersection-over-Union (IoU), single-frame Segmentation and Tracking Quality ($\mathrm{STQ}_1$), and single-frame Association Quality ($\mathrm{AQ}_1$)~\cite{luz2026lags}.
Temporal consistency assesses tracking quality and identity continuity using Segmentation and Tracking Quality (STQ) and Association Quality (AQ)~\cite{weber2021step,chen2025trackocc}, Higher Order Tracking Accuracy (HOTA)~\cite{luiten2021hota}, and Identification F1 Score (IDF1)~\cite{ristani2016identity}.
Asset quality uses framewise 3D oriented bounding-box Average Precision (AP) against annotated object boxes within the evaluation region, reported at IoU thresholds of 0.25 ($\mathrm{AP}_{25}$) and 0.50 ($\mathrm{AP}_{50}$).
Asset Fragmentation (AF) measures the mean number of additional persistent asset records per GT object matched at least once within a continuous evaluation segment.

\begin{figure*}[t]
    \centering
    \includegraphics[width=\linewidth]{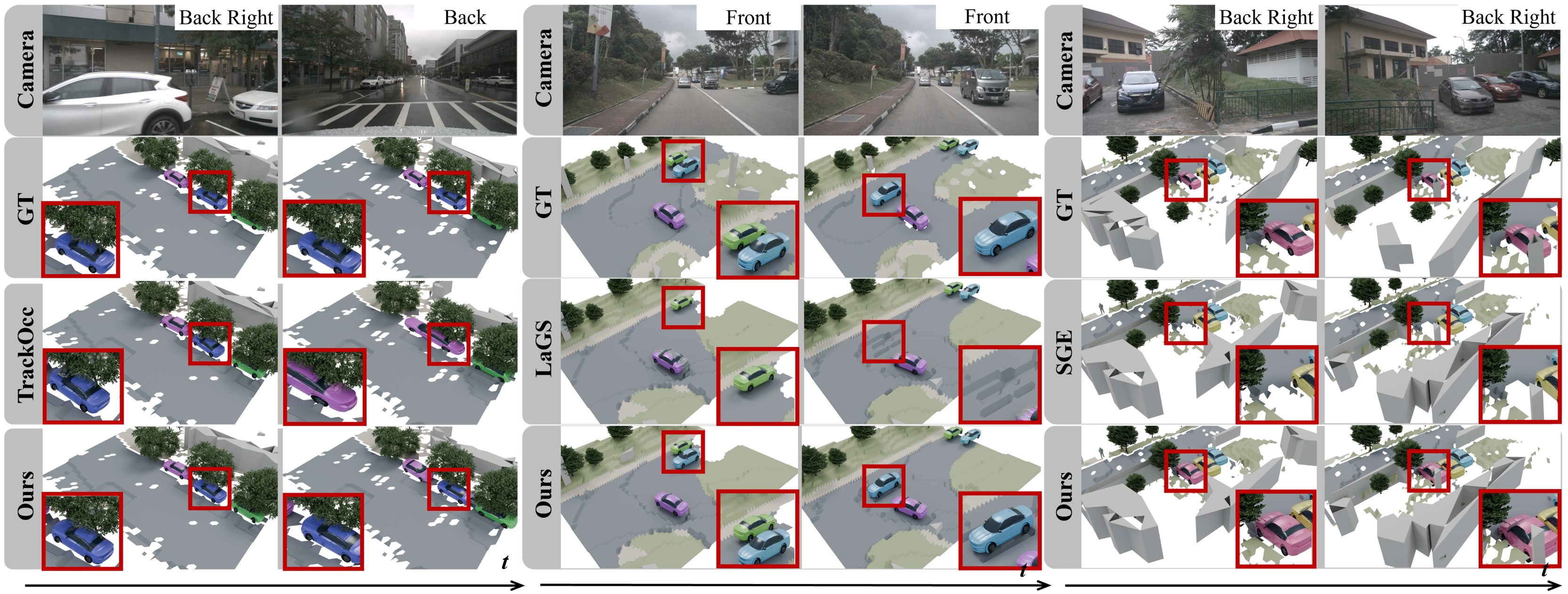}
    \vspace{-1.5em}
    \caption{\textbf{Qualitative comparison of street-scene assets.} Each group compares a baseline with Video2Asset (Ours) built on the corresponding occupancy model: TrackOcc (left), LaGS (middle), and SGE (right).  GT denotes a reference scene instantiated from ground-truth annotations.}
    \vspace{-1em}
    \label{fig:comparison}
\end{figure*} 

\subsection{Comparison}
\label{sec:experiments:comparison}
\begin{table*}[t]
    \centering
    \caption{\textbf{Ablation experiments.} Results on CarlaOcc with TrackOcc as the base model.}
    \vspace{-1em}
    \label{tab:exp:ablation}
    \small
    \setlength{\tabcolsep}{2.5pt}
    \renewcommand{\arraystretch}{1.08}
    \begin{tabular*}{\textwidth}{@{\extracolsep{\fill}}lcc*{11}{r}@{}}
        \toprule
        \multirow[c]{2}{*}{Variant} & \multirow[c]{2}{*}{HCG} & \multirow[c]{2}{*}{AEA} & \multicolumn{4}{c}{Observation Quality} & \multicolumn{4}{c}{Temporal Metrics} & \multicolumn{3}{c}{Asset Quality} \\
        \cmidrule(lr){4-7} \cmidrule(lr){8-11} \cmidrule(lr){12-14}
        & & & mIoU $\uparrow$ & $\mathrm{STQ}_1\uparrow$ & $\mathrm{AQ}_1\uparrow$ & IoU $\uparrow$ & STQ $\uparrow$ & AQ $\uparrow$ & HOTA $\uparrow$ & IDF1 $\uparrow$ & $\mathrm{AP}_{25}\uparrow$ & $\mathrm{AP}_{50}\uparrow$ & $\mathrm{AF}\downarrow$ \\
        \midrule
        Base & $\times$ & $\times$ & 33.92 & 25.91 & 19.80 & 56.91 & 12.89 & 4.90 & 11.25 & 7.24 & 8.88 & 6.62 & 6.35 \\
        + HCG & $\checkmark$ & $\times$ & 35.73 & 29.17 & 23.83 & 57.08 & 18.02 & 9.09 & 17.68 & 13.78 & 19.26 & 15.37 & 4.83 \\
        + AEA & $\times$ & $\checkmark$ & 33.92 & 25.91 & 19.80 & 56.91 & 20.20 & 12.03 & 22.81 & 20.08 & 8.88 & 6.62 & 3.67 \\
        Full & $\checkmark$ & $\checkmark$ & \textbf{35.73} & \textbf{29.17} & \textbf{23.83} & \textbf{57.08} & \textbf{23.24} & \textbf{15.12} & \textbf{25.99} & \textbf{24.59} & \textbf{19.29} & \textbf{15.45} & \textbf{2.75} \\
        \bottomrule
    \end{tabular*}
    \vspace{-1em}
\end{table*}
\textbf{Quantitative Comparisons.}
For observation quality, Video2Asset improves single-frame instance grouping across TrackOcc~\cite{chen2025trackocc}, LaGS~\cite{luz2026lags}, and SGE~\cite{luz2026sge} on all three datasets (Table~\ref{tab:exp:comparison}).
For LaGS on CarlaOcc, $\mathrm{STQ}_1$ and $\mathrm{AQ}_1$ increase by \textbf{0.71} and \textbf{1.20} points, respectively.
SGE also improves $\mathrm{AQ}_1$ from 22.99 to \textbf{23.25} on Waymo, alongside improved mIoU and occupancy IoU.
These gains indicate more accurate voxel-to-instance assignments, providing cleaner object-specific geometry and appearance evidence for subsequent temporal association.

For temporal metrics, all four scores improve across the nine paired comparisons, indicating more consistent object identities over time.
TrackOcc shows the largest gains on CarlaOcc, with HOTA rising from 11.25 to \textbf{25.99} and IDF1 from 7.24 to \textbf{24.59}.
LaGS and SGE also improve IDF1 on CarlaOcc by \textbf{0.24} and \textbf{0.28} points, respectively.
These gains reflect improved cross-frame identity continuity, supporting more consistent observation histories for persistent asset construction.

Video2Asset also improves asset quality while reducing identity fragmentation.
On CarlaOcc, LaGS gains \textbf{1.50} and \textbf{1.14} points in $\mathrm{AP}_{25}$ and $\mathrm{AP}_{50}$, respectively, while SGE raises $\mathrm{AP}_{50}$ from 26.31 to \textbf{27.29}.
SGE also gains \textbf{0.68} points in $\mathrm{AP}_{25}$ on Waymo, extending the asset-level benefit to real-world scenes.
AF falls from 6.35 to \textbf{2.75} for TrackOcc on CarlaOcc; on nuScenes, LaGS and SGE reduce AF from 0.26 to \textbf{0.24} and \textbf{0.22}, respectively.
Higher AP and lower AF indicate more accurate spatial alignment and fewer distinct asset records representing the same physical object.

For nuScenes GT objects that remain within the evaluation region and experience a continuous low-visibility interval (0--40\%) longer than 2\,s between high-visibility observations (80--100\%), we compute AF over their complete scene trajectories. Table~\ref{tab:exp:low-visibility} shows that Video2Asset reduces AF from 2.48 to \textbf{1.90} on TrackOcc, and by \textbf{0.18} and \textbf{0.09} on LaGS and SGE, respectively.
These results show that Video2Asset better preserves asset identities on this challenging subset.

\textbf{Qualitative Comparisons.}
For Figure~\ref{fig:comparison}, we export the panoptic occupancy predictions of TrackOcc, LaGS, and SGE into asset scenes using our mesh-replacement procedure, pairing each baseline with Video2Asset built on the same occupancy model. At the second timestamp, TrackOcc merges two vehicle identities into one elongated mesh, distorting the scene’s object composition and geometry, whereas Video2Asset preserves two separate assets across both timestamps. In the LaGS and SGE groups, the highlighted vehicles are missing from the baseline-derived scenes at both timestamps but remain present in the corresponding Video2Asset results. These comparisons show better object separation and identity continuity, together with more complete asset scenes.

\subsection{Evaluations}
\label{sec:experiments:evaluations}

\textbf{Ablation Studies.}
HCG and AEA make complementary contributions to asset construction (Table~\ref{tab:exp:ablation}).
HCG improves current-frame instance grouping and box-level asset quality, increasing $\mathrm{AQ}_1$ by \textbf{4.03} points and $\mathrm{AP}_{50}$ by \textbf{8.75} points over the base model.
AEA strengthens identity continuity, improving HOTA and IDF1 by \textbf{11.56} and \textbf{12.84} points, respectively.
AF drops from 6.35 to \textbf{4.83} with HCG and to \textbf{3.67} with AEA, showing that improved instance evidence and association both reduce fragmentation of asset records.
Combining both components yields the highest temporal and box-level AP scores, with AF falling further to \textbf{2.75}.
The full model further improves IDF1 by \textbf{4.51} points over AEA alone and \textbf{10.81} points over HCG alone.

Table~\ref{tab:exp:hcg} compares full HCG with native TrackOcc (Base), parameter-matched current-frame local guidance (Local), and shared historical guidance (Shared).
Local samples current features near each query reference point and improves IDF1 over Base by \textbf{1.56} points.
Shared pools current features from pose- and motion-aligned historical instance regions into a common summary for every query, but lowers $\mathrm{AQ}_1$ from 12.50 (Base) to 11.00, indicating that mixing support across instances can impair grouping.
Full HCG retains instance-specific historical guidance and improves $\mathrm{AQ}_1$ and IDF1 over Local by \textbf{2.77} and \textbf{6.27}, respectively.

\begin{table}[t]
    \centering
    \caption{\textbf{HCG ablation experiments.} Results on nuScenes with TrackOcc as the base model.}
    \vspace{-1em}
    \label{tab:exp:hcg}
    \small
    \setlength{\tabcolsep}{4pt}
    \renewcommand{\arraystretch}{1.08}
    \begin{tabular*}{\columnwidth}{@{\extracolsep{\fill}}lrrrr@{}}
        \toprule
        Metric $\uparrow$ & Base & Local & Shared & HCG \\
        \midrule
        mIoU & 33.64 & 33.25 & 32.25 & \textbf{33.85} \\
        $\mathrm{STQ}_1$ & 20.51 & 20.48 & 18.84 & \textbf{22.82} \\
        $\mathrm{AQ}_1$ & 12.50 & 12.61 & 11.00 & \textbf{15.38} \\
        IoU & 63.37 & 63.62 & 63.54 & \textbf{63.69} \\
        \midrule
        STQ & 13.14 & 13.65 & 12.43 & \textbf{16.34} \\
        AQ & 5.14 & 5.60 & 4.79 & \textbf{7.89} \\
        HOTA & 13.58 & 14.71 & 13.01 & \textbf{19.97} \\
        IDF1 & 10.76 & 12.32 & 10.45 & \textbf{18.59} \\
        \bottomrule
    \end{tabular*}
    \vspace{-1em}
\end{table}

Table~\ref{tab:exp:association} evaluates multi-frame evidence accumulation and selective history reading in AEA with HCG observations fixed.
The geometric tracker uses Kalman prediction and 3D GIoU-based Hungarian matching on mask-derived world-frame proxy boxes; simple memory association instead uses motion-gated cosine matching of mean identity and appearance descriptors.
Simple memory improves over geometric tracking, while full AEA further raises HOTA and IDF1 by \textbf{0.88} and \textbf{1.28} points.
Decoupling candidate ranking from $\mathrm{Birth}/\mathrm{Inherit}$ provides additional HOTA and IDF1 gains of \textbf{0.14} and \textbf{0.47} points, respectively.
Single-frame history and uniform aggregation lower IDF1 by \textbf{2.67} and \textbf{2.49} points, respectively, highlighting the benefits of evidence accumulation and selective history reading.
Removing any evidence branch also lowers all four temporal metrics, with the largest IDF1 drop (\textbf{1.54} points) occurring without identity evidence.

\begin{table}[t]
    \centering
    \caption{\textbf{AEA ablation experiments.} Results on nuScenes with TrackOcc as the base model.}
    \vspace{-1em}
    \label{tab:exp:association}
    \small
    \setlength{\tabcolsep}{1.5pt}
    \renewcommand{\arraystretch}{1.08}
    \begin{tabular*}{\columnwidth}{@{\extracolsep{\fill}}lrrrr@{}}
        \toprule
        Variant & STQ $\uparrow$ & AQ $\uparrow$ & HOTA $\uparrow$ & IDF1 $\uparrow$ \\
        \midrule
        HCG only & 16.34 & 7.89 & 19.97 & 18.59 \\
        Geometry tracker & 16.13 & 7.69 & 19.23 & 18.49 \\
        Simple memory association &  16.66 & 8.20 & 21.35 & 20.03 \\
        \midrule
        w/o decoupling & 16.93 & 8.47 & 22.09 & 20.84 \\
        Single-frame history & 16.36 & 7.91 & 19.95 & 18.64 \\
        Uniform history aggregation & 16.38 & 7.93 & 20.14 & 18.82 \\
        w/o identity & 16.44 & 7.98 & 20.36 & 19.77 \\
        w/o appearance & 16.88 & 8.42 & 21.27 & 20.05 \\
        w/o geometry & 16.72 & 8.26 & 21.25 & 20.00 \\
        Full AEA & \textbf{17.12} & \textbf{8.66} & \textbf{22.23} & \textbf{21.31} \\
        \bottomrule
    \end{tabular*}
    \vspace{-1em}
\end{table}

\begin{figure}[t]
    \centering
    \includegraphics[width=\linewidth]{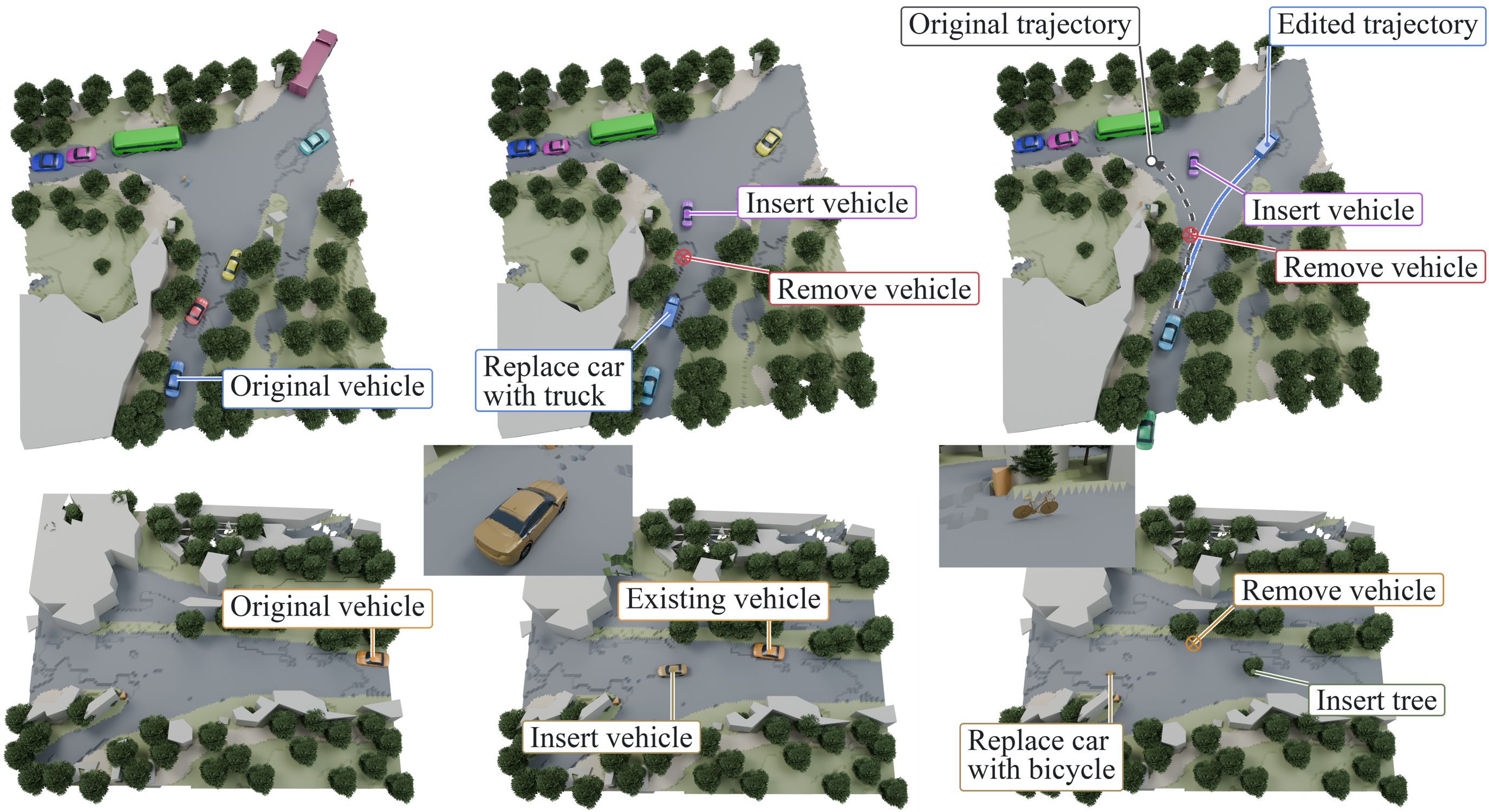}
    \vspace{-1.5em}
    \caption{Asset-level editing of exported street scenes.}
    \vspace{-1em}
    \label{fig:exp:operations}
\end{figure}

\textbf{Asset Operations.}
Persistent object records enable asset-level editing of the exported scene (Figure~\ref{fig:exp:operations}).
The top row shows vehicle insertion and removal, car-to-truck replacement, and trajectory editing.
The bottom row replaces an inserted car with a bicycle, removes an existing vehicle, and adds a tree.
These edits modify object presence, geometry, and motion while preserving the surrounding road and buildings.

\textbf{Efficiency.}
Video2Asset introduces persistent asset modeling through lightweight identity encoding and association, with additional computation devoted to identity-aware observation construction, history reading, and state maintenance.
As shown in Table~\ref{tab:exp:efficiency}, we measure per-frame latency on nuScenes.
Across the three variants, Video2Asset adds 16.41--18.81\,ms per frame relative to the corresponding baselines.
The modest overhead reflects the lightweight design of our persistent asset modeling.

\begin{table}[t]
    \centering
    \caption{\textbf{Runtime breakdown.} Latency (ms/frame) on nuScenes.}
    \vspace{-1em}
    \label{tab:exp:efficiency}
    \small
    \setlength{\tabcolsep}{4pt}
    \renewcommand{\arraystretch}{1.08}
    \begin{tabular*}{\columnwidth}{@{\extracolsep{\fill}}lrrr@{}}
        \toprule
        \multirow[c]{2}{*}{Component} & Ours & Ours & Ours \vspace{-0.3em}\\
        & {\scriptsize (with TrackOcc)} & {\scriptsize (with LaGS)} & {\scriptsize (with SGE)} \\
        \midrule
        Baseline & 127.32 & 303.08 & 324.99 \\
        HCG & 3.67 & 5.89 & 4.24 \\
        Identity encoder & 4.30 & 4.46 & 3.17 \\
        AEA & 8.44 & 8.46 & 9.02 \\
        \midrule
        Total & 143.73 & 321.89 & 341.42 \\
        \bottomrule
    \end{tabular*}
    \vspace{-1em}
\end{table}

\section{Discussion}
\label{sec:discussion}

\textbf{Limitations.}
Although Video2Asset supports persistent, editable street-scene construction, two limitations remain.
First, class-selected library meshes do not fully capture instance-specific shape details.
Second, assets support object-level poses and trajectories, with articulated motion limited to predefined animations.

\textbf{Conclusion.}
We presented Video2Asset, an online framework for constructing persistent, editable 3D street-scene assets from multi-view video.
Using occupancy as an intermediate representation, HCG guides instance decoding and an identity encoder fuses 3D features with appearance to construct identity-aware observations.
AEA compares observations with evidence accumulated within asset records to determine identity inheritance and update world-coordinate states for mesh instantiation and editing.
Multi-dataset experiments show improved observation quality, temporal consistency, and asset quality.

{
    \small
    \bibliographystyle{ieeenat_fullname}
    \bibliography{main}

@String(ECCV= {Eur. Conf. Comput. Vis.})

@String(AAAI = {AAAI})

@String(ECCV  = {ECCV})

@misc{chen2025trackocc,
  author = {Zhuoguang Chen and Kenan Li and Xiuyu Yang and Tao Jiang and Yiming Li and Hang Zhao},
  title = {{TrackOcc}: Camera-based {4D} Panoptic Occupancy Tracking},
  year = {2025},
  howpublished = {arXiv:2503.08471},
  eprint = {2503.08471},
  archivePrefix = {arXiv}
}

@inproceedings{wei2023surroundocc,
  author = {Yi Wei and Linqing Zhao and Wenzhao Zheng and Zheng Zhu and Jie Zhou and Jiwen Lu},
  title = {{SurroundOcc}: Multi-camera {3D} Occupancy Prediction for Autonomous Driving},
  booktitle = {Proceedings of the IEEE/CVF International Conference on Computer Vision},
  year = {2023},
  pages = {21729--21740}
}

@inproceedings{zhang2023occformer,
  author = {Yunpeng Zhang and Zheng Zhu and Dalong Du},
  title = {{OccFormer}: Dual-path Transformer for Vision-based {3D} Semantic Occupancy Prediction},
  booktitle = {Proceedings of the IEEE/CVF International Conference on Computer Vision},
  year = {2023},
  pages = {9433--9443}
}

@inproceedings{huang2023tpvformer,
  author = {Yuanhui Huang and Wenzhao Zheng and Yunpeng Zhang and Jie Zhou and Jiwen Lu},
  title = {Tri-Perspective View for Vision-Based {3D} Semantic Occupancy Prediction},
  booktitle = {Proceedings of the IEEE/CVF Conference on Computer Vision and Pattern Recognition},
  year = {2023},
  pages = {9223--9232}
}

@inproceedings{wang2024panoocc,
  author = {Yuqi Wang and Yuntao Chen and Xingyu Liao and Lue Fan and Zhaoxiang Zhang},
  title = {{PanoOcc}: Unified Occupancy Representation for Camera-based {3D} Panoptic Segmentation},
  booktitle = {Proceedings of the IEEE/CVF Conference on Computer Vision and Pattern Recognition},
  year = {2024},
  pages = {17158--17168}
}

@inproceedings{liu2024sparseocc,
  author = {Haisong Liu and Yang Chen and Haiguang Wang and Zetong Yang and Tianyu Li and Jia Zeng and Li Chen and Hongyang Li and Limin Wang},
  title = {Fully Sparse {3D} Occupancy Prediction},
  booktitle = {European Conference on Computer Vision},
  year = {2024}
}

@article{luz2026lags,
  author = {Maximilian Luz and Rohit Mohan and Thomas N{\"u}rnberg and Yakov Miron and Daniele Cattaneo and Abhinav Valada},
  title = {Latent Gaussian Splatting for {4D} Panoptic Occupancy Tracking},
  journal = {IEEE Robotics and Automation Letters},
  year = {2026},
  volume = {11},
  number = {8},
  doi = {10.1109/LRA.2026.3703990}
}

@inproceedings{zuo2025gaussianworld,
  author = {Sicheng Zuo and Wenzhao Zheng and Yuanhui Huang and Jie Zhou and Jiwen Lu},
  title = {{GaussianWorld}: Gaussian World Model for Streaming {3D} Occupancy Prediction},
  booktitle = {Proceedings of the IEEE/CVF Conference on Computer Vision and Pattern Recognition},
  year = {2025},
  pages = {6772--6781}
}

@inproceedings{luz2026sge,
  author = {Maximilian Luz and Thomas N{\"u}rnberg and Yakov Miron and Abhinav Valada},
  title = {Streaming Gaussian Encoding for {4D} Panoptic Occupancy Tracking},
  booktitle = {IEEE/RSJ International Conference on Intelligent Robots and Systems},
  year = {2026}
}

@inproceedings{ost2021neuralscenegraphs,
  author = {Julian Ost and Fahim Mannan and Nils Thuerey and Julian Knodt and Felix Heide},
  title = {Neural Scene Graphs for Dynamic Scenes},
  booktitle = {Proceedings of the IEEE/CVF Conference on Computer Vision and Pattern Recognition},
  year = {2021},
  pages = {2856--2865}
}

@inproceedings{yan2024street,
  author = {Yunzhi Yan and Haotong Lin and Chenxu Zhou and Weijie Wang and Haiyang Sun and Kun Zhan and Xianpeng Lang and Xiaowei Zhou and Sida Peng},
  title = {{Street Gaussians} for Modeling Dynamic Urban Scenes},
  booktitle = {European Conference on Computer Vision},
  year = {2024}
}

@inproceedings{chen2025omnire,
  author = {Ziyu Chen and Jiawei Yang and Jiahui Huang and Riccardo de Lutio and Janick Martinez Esturo and Boris Ivanovic and Or Litany and Zan Gojcic and Sanja Fidler and Marco Pavone and Li Song and Yue Wang},
  title = {{OmniRe}: Omni Urban Scene Reconstruction},
  booktitle = {International Conference on Learning Representations},
  year = {2025}
}

@misc{zhou2024hugsim,
  author = {Hongyu Zhou and Longzhong Lin and Jiabao Wang and Yichong Lu and Dongfeng Bai and Bingbing Liu and Yue Wang and Andreas Geiger and Yiyi Liao},
  title = {{HUGSIM}: A Real-Time, Photo-Realistic and Closed-Loop Simulator for Autonomous Driving},
  year = {2024},
  howpublished = {arXiv:2412.01718},
  eprint = {2412.01718},
  archivePrefix = {arXiv}
}

@inproceedings{mildenhall2020nerf,
  author = {Ben Mildenhall and Pratul P. Srinivasan and Matthew Tancik and Jonathan T. Barron and Ravi Ramamoorthi and Ren Ng},
  title = {{NeRF}: Representing Scenes as Neural Radiance Fields for View Synthesis},
  booktitle = {European Conference on Computer Vision},
  year = {2020}
}

@article{kerbl2023gaussiansplatting,
  author = {Bernhard Kerbl and Georgios Kopanas and Thomas Leimk{\"u}hler and George Drettakis},
  title = {{3D} Gaussian Splatting for Real-Time Radiance Field Rendering},
  journal = {ACM Transactions on Graphics},
  volume = {42},
  number = {4},
  year = {2023}
}

@inproceedings{niemeyer2022regnerf,
  author = {Michael Niemeyer and Jonathan T. Barron and Ben Mildenhall and Mehdi S. M. Sajjadi and Andreas Geiger and Noha Radwan},
  title = {{RegNeRF}: Regularizing Neural Radiance Fields for View Synthesis From Sparse Inputs},
  booktitle = {Proceedings of the IEEE/CVF Conference on Computer Vision and Pattern Recognition},
  year = {2022}
}

@inproceedings{zhu2024fsgs,
  author = {Zehao Zhu and Zhiwen Fan and Yifan Jiang and Zhangyang Wang},
  title = {{FSGS}: Real-Time Few-shot View Synthesis using Gaussian Splatting},
  booktitle = {European Conference on Computer Vision},
  year = {2024}
}

@inproceedings{yao2018mvsnet,
  author = {Yao Yao and Zixin Luo and Shiwei Li and Tian Fang and Long Quan},
  title = {{MVSNet}: Depth Inference for Unstructured Multi-view Stereo},
  booktitle = {European Conference on Computer Vision},
  year = {2018}
}

@inproceedings{schmid2024khronos,
  author = {Lukas Schmid and Marcus Abate and Yun Chang and Luca Carlone},
  title = {{Khronos}: A Unified Approach for Spatio-Temporal Metric-Semantic {SLAM} in Dynamic Environments},
  booktitle = {Proceedings of Robotics: Science and Systems},
  year = {2024},
  doi = {10.15607/RSS.2024.XX.081}
}

@misc{yu2026superocc,
  author = {Zichen Yu and Quanli Liu and Wei Wang and Liyong Zhang and Xiaoguang Zhao},
  title = {{SuperOcc}: Toward Cohesive Temporal Modeling for Superquadric-based {3D} Occupancy Prediction},
  year = {2026},
  howpublished = {arXiv:2601.15644},
  eprint = {2601.15644},
  archivePrefix = {arXiv}
}

@inproceedings{cai2022memot,
  author = {Jiarui Cai and Mingze Xu and Wei Li and Yuanjun Xiong and Wei Xia and Zhuowen Tu and Stefano Soatto},
  title = {{MeMOT}: Multi-Object Tracking With Memory},
  booktitle = {Proceedings of the IEEE/CVF Conference on Computer Vision and Pattern Recognition},
  year = {2022},
  pages = {8090--8100}
}

@inproceedings{gao2023memotr,
  author = {Ruopeng Gao and Limin Wang},
  title = {{MeMOTR}: Long-Term Memory-Augmented Transformer for Multi-Object Tracking},
  booktitle = {Proceedings of the IEEE/CVF International Conference on Computer Vision},
  year = {2023},
  pages = {9901--9910}
}

@inproceedings{lee2025lomm,
  author = {Seunghun Lee and Jiwan Seo and Minwoo Choi and Kiljoon Han and Jahoon Jeong and Zane Durante and Ehsan Adeli and Sang Hyun Park and Sunghoon Im},
  title = {{LOMM}: Latest Object Memory Management for Temporally Consistent Video Instance Segmentation},
  booktitle = {Proceedings of the IEEE/CVF International Conference on Computer Vision},
  year = {2025},
  pages = {13719--13729}
}

@inproceedings{zhong2025prismgs,
  author = {Houqiang Zhong and Zhenglong Wu and Sihua Fu and Zihan Zheng and Xin Jin and Xiaoyun Zhang and Li Song and Qiang Hu},
  title = {{PrismGS}: Physically-Grounded Anti-Aliasing for High-Fidelity Large-Scale {3D} Gaussian Splatting},
  booktitle = {International Conference on Visual Communications and Image Processing},
  year = {2025},
  pages = {1--5},
  doi = {10.1109/VCIP67698.2025.11396821}
}

@misc{hu2026gensplatcodec,
  author = {Qiang Hu and Zhenlong Wu and Lei Huang and Zihan Zheng and Xiaoyun Zhang and Wenjun Zhang},
  title = {{GenSplatCodec}: Feed-Forward Gaussian Splatting Compression via One-Step Diffusion},
  year = {2026},
  howpublished = {arXiv:2607.24403},
  eprint = {2607.24403},
  archivePrefix = {arXiv}
}

@inproceedings{wang2021neus,
  author = {Peng Wang and Lingjie Liu and Yuan Liu and Christian Theobalt and Taku Komura and Wenping Wang},
  title = {{NeuS}: Learning Neural Implicit Surfaces by Volume Rendering for Multi-view Reconstruction},
  booktitle = {Advances in Neural Information Processing Systems},
  volume = {34},
  year = {2021},
  pages = {27171--27183}
}

@inproceedings{li2023neuralangelo,
  author = {Zhaoshuo Li and Thomas M{\"u}ller and Alex Evans and Russell H. Taylor and Mathias Unberath and Ming-Yu Liu and Chen-Hsuan Lin},
  title = {{Neuralangelo}: High-Fidelity Neural Surface Reconstruction},
  booktitle = {Proceedings of the IEEE/CVF Conference on Computer Vision and Pattern Recognition},
  year = {2023},
  pages = {8456--8465}
}

@inproceedings{zheng2024jointrf,
  author = {Zihan Zheng and Houqiang Zhong and Qiang Hu and Xiaoyun Zhang and Li Song and Ya Zhang and Yanfeng Wang},
  title = {{JointRF}: End-to-End Joint Optimization for Dynamic Neural Radiance Field Representation and Compression},
  booktitle = {IEEE International Conference on Image Processing},
  year = {2024},
  pages = {3292--3298},
  doi = {10.1109/ICIP51287.2024.10647336}
}

@inproceedings{zheng2024hpc,
  author = {Zihan Zheng and Houqiang Zhong and Qiang Hu and Xiaoyun Zhang and Li Song and Ya Zhang and Yanfeng Wang},
  title = {{HPC}: Hierarchical Progressive Coding Framework for Volumetric Video},
  booktitle = {Proceedings of the 32nd ACM International Conference on Multimedia},
  year = {2024},
  pages = {7937--7946},
  doi = {10.1145/3664647.3681107}
}

@inproceedings{hu2025vrvvc,
  author = {Qiang Hu and Houqiang Zhong and Zihan Zheng and Xiaoyun Zhang and Zhengxue Cheng and Li Song and Guangtao Zhai and Yanfeng Wang},
  title = {{VRVVC}: Variable-Rate {NeRF}-Based Volumetric Video Compression},
  booktitle = {Proceedings of the AAAI Conference on Artificial Intelligence},
  volume = {39},
  number = {4},
  year = {2025},
  pages = {3563--3571},
  doi = {10.1609/aaai.v39i4.32370}
}

@inproceedings{hu2025_4dgc,
  author = {Qiang Hu and Zihan Zheng and Houqiang Zhong and Sihua Fu and Li Song and Xiaoyun Zhang and Guangtao Zhai and Yanfeng Wang},
  title = {{4DGC}: Rate-Aware {4D} Gaussian Compression for Efficient Streamable Free-Viewpoint Video},
  booktitle = {Proceedings of the IEEE/CVF Conference on Computer Vision and Pattern Recognition},
  year = {2025},
  pages = {875--885}
}

@inproceedings{zheng2025_4dgcpro,
  author = {Zihan Zheng and Zhenlong Wu and Houqiang Zhong and Yuan Tian and Ning Cao and Lan Xu and Jiangchao Yao and Xiaoyun Zhang and Qiang Hu and Wenjun Zhang},
  title = {{4DGCPro}: Efficient Hierarchical {4D} Gaussian Compression for Progressive Volumetric Video Streaming},
  booktitle = {Advances in Neural Information Processing Systems},
  volume = {38},
  year = {2025}
}

@article{zheng2026cdgs,
  author = {Zihan Zheng and Zhenlong Wu and Xuanxuan Wang and Houqiang Zhong and Xiaoyun Zhang and Qiang Hu and Guangtao Zhai and Wenjun Zhang},
  title = {Constrained Dynamic Gaussian Splatting},
  journal = {IEEE Transactions on Circuits and Systems for Video Technology},
  year = {2026},
  doi = {10.1109/TCSVT.2026.3740925}
}

@inproceedings{cao2022monoscene,
  author = {Anh-Quan Cao and Raoul de Charette},
  title = {{MonoScene}: Monocular {3D} Semantic Scene Completion},
  booktitle = {Proceedings of the IEEE/CVF Conference on Computer Vision and Pattern Recognition},
  year = {2022}
}

@inproceedings{li2023voxformer,
  author = {Yiming Li and Zhiding Yu and Christopher Choy and Chaowei Xiao and Jose M. Alvarez and Sanja Fidler and Chen Feng and Anima Anandkumar},
  title = {{VoxFormer}: Sparse Voxel Transformer for Camera-Based {3D} Semantic Scene Completion},
  booktitle = {Proceedings of the IEEE/CVF Conference on Computer Vision and Pattern Recognition},
  year = {2023},
  pages = {9087--9098}
}

@inproceedings{huang2024gaussianformer,
  author = {Yuanhui Huang and Wenzhao Zheng and Yunpeng Zhang and Jie Zhou and Jiwen Lu},
  title = {{GaussianFormer}: Scene as Gaussians for Vision-Based {3D} Semantic Occupancy Prediction},
  booktitle = {European Conference on Computer Vision},
  year = {2024}
}

@inproceedings{leng2025stocc,
  author = {Ziyang Leng and Jiawei Yang and Wenlong Yi and Bolei Zhou},
  title = {Occupancy Learning with Spatiotemporal Memory},
  booktitle = {Proceedings of the IEEE/CVF International Conference on Computer Vision},
  year = {2025},
  pages = {26569--26578}
}

@misc{feng2026carlaocc,
  author = {Yi Feng and Junwu E and Zizhan Guo and Yu Ma and Hanli Wang and Rui Fan},
  title = {An Instance-Centric Panoptic Occupancy Prediction Benchmark for Autonomous Driving},
  year = {2026},
  howpublished = {arXiv:2603.27238},
  eprint = {2603.27238},
  archivePrefix = {arXiv}
}

@inproceedings{caesar2020nuscenes,
  author = {Holger Caesar and Varun Bankiti and Alex H. Lang and Sourabh Vora and Venice Erin Liong and Qiang Xu and Anush Krishnan and Yu Pan and Giancarlo Baldan and Oscar Beijbom},
  title = {{nuScenes}: A Multimodal Dataset for Autonomous Driving},
  booktitle = {Proceedings of the IEEE/CVF Conference on Computer Vision and Pattern Recognition},
  year = {2020}
}

@inproceedings{sun2020waymo,
  author = {Pei Sun and Henrik Kretzschmar and Xerxes Dotiwalla and Aurelien Chouard and Vijaysai Patnaik and Paul Tsui and James Guo and Yin Zhou and Yuning Chai and Benjamin Caine and Vijay Vasudevan and Wei Han and Jiquan Ngiam and Hang Zhao and Aleksei Timofeev and Scott Ettinger and Maxim Krivokon and Amy Gao and Aditya Joshi and Yu Zhang and Jonathon Shlens and Zhifeng Chen and Dragomir Anguelov},
  title = {Scalability in Perception for Autonomous Driving: {Waymo} Open Dataset},
  booktitle = {Proceedings of the IEEE/CVF Conference on Computer Vision and Pattern Recognition},
  year = {2020}
}

@inproceedings{tian2023occ3d,
  author = {Xiaoyu Tian and Tao Jiang and Longfei Yun and Yucheng Mao and Huitong Yang and Yue Wang and Yilun Wang and Hang Zhao},
  title = {{Occ3D}: A Large-Scale {3D} Occupancy Prediction Benchmark for Autonomous Driving},
  booktitle = {Advances in Neural Information Processing Systems},
  volume = {36},
  pages = {64318--64330},
  year = {2023}
}

@inproceedings{weber2021step,
  author = {Mark Weber and Jun Xie and Maxwell Collins and Yukun Zhu and Paul Voigtlaender and Hartwig Adam and Bradley Green and Andreas Geiger and Bastian Leibe and Daniel Cremers and Aljo{\v{s}}a O{\v{s}}ep and Laura Leal-Taix{\'e} and Liang-Chieh Chen},
  title = {{STEP}: Segmenting and Tracking Every Pixel},
  booktitle = {Proceedings of the Neural Information Processing Systems Track on Datasets and Benchmarks},
  volume = {1},
  year = {2021}
}

@article{luiten2021hota,
  author = {Jonathon Luiten and Aljo{\v{s}}a O{\v{s}}ep and Patrick Dendorfer and Philip Torr and Andreas Geiger and Laura Leal-Taix{\'e} and Bastian Leibe},
  title = {{HOTA}: A Higher Order Metric for Evaluating Multi-Object Tracking},
  journal = {International Journal of Computer Vision},
  volume = {129},
  number = {2},
  pages = {548--578},
  year = {2021},
  doi = {10.1007/s11263-020-01375-2}
}

@inproceedings{ristani2016identity,
  author = {Ergys Ristani and Francesco Solera and Roger S. Zou and Rita Cucchiara and Carlo Tomasi},
  title = {Performance Measures and a Data Set for Multi-Target, Multi-Camera Tracking},
  booktitle = {Computer Vision -- ECCV 2016 Workshops},
  pages = {17--35},
  year = {2016},
  doi = {10.1007/978-3-319-48881-3_2}
}
}


\end{document}